\documentclass[10pt,twocolumn,letterpaper]{article}

\usepackage[pagenumbers]{cvpr} 

\usepackage[dvipsnames]{xcolor}

\newcommand{\cM}{\mathcal{M}}

\newcommand{\cV}{\mathcal{V}}

\newcommand{\cZ}{\mathcal{Z}}

\newcommand{\figref}[1]{Fig.~\ref{#1}}

\makeatletter
\DeclareRobustCommand\onedot{\futurelet\@let@token\@onedot}
\def\@onedot{\ifx\@let@token.\else.\null\fi\xspace}
\def\eg{e.g\onedot} 
\def\ie{i.e\onedot} 
 
\def\etc{etc\onedot}

\def\etal{et~al\onedot}

\makeatother

\newcommand{\PAR}[1]{\vspace{0.1cm}\noindent{\bf #1} }

\definecolor{cvprblue}{rgb}{0.21,0.49,0.74}
\usepackage[pagebackref,breaklinks,colorlinks,citecolor=cvprblue]{hyperref}

\usepackage{wrapfig}

\def\paperID{272} 
\def\confName{3DV\xspace}
\def\confYear{2025\xspace}

\title{Controllable Text-to-3D Generation via Surface-Aligned Gaussian Splatting}

\author{Zhiqi Li$^{1,2}$\qquad Yiming Chen$^{1,2}$ \qquad Lingzhe Zhao$^{2}$ \qquad Peidong Liu$^{2,\dag}$\vspace{0.1cm} \\
 $^{1}$Zhejiang University 
 \qquad 
 $^{2}$Westlake University \\
{\tt\small \{lizhiqi49, chenyiming, zhaolingzhe, liupeidong\}@westlake.edu.cn}
}

\begin{document}
\maketitle
\let\thefootnote\relax\footnotetext{$^{\dag}$ Corresponding author.}

\begin{figure*}\label{fig_teaser}
	\centering
	\includegraphics[width=0.95\linewidth]{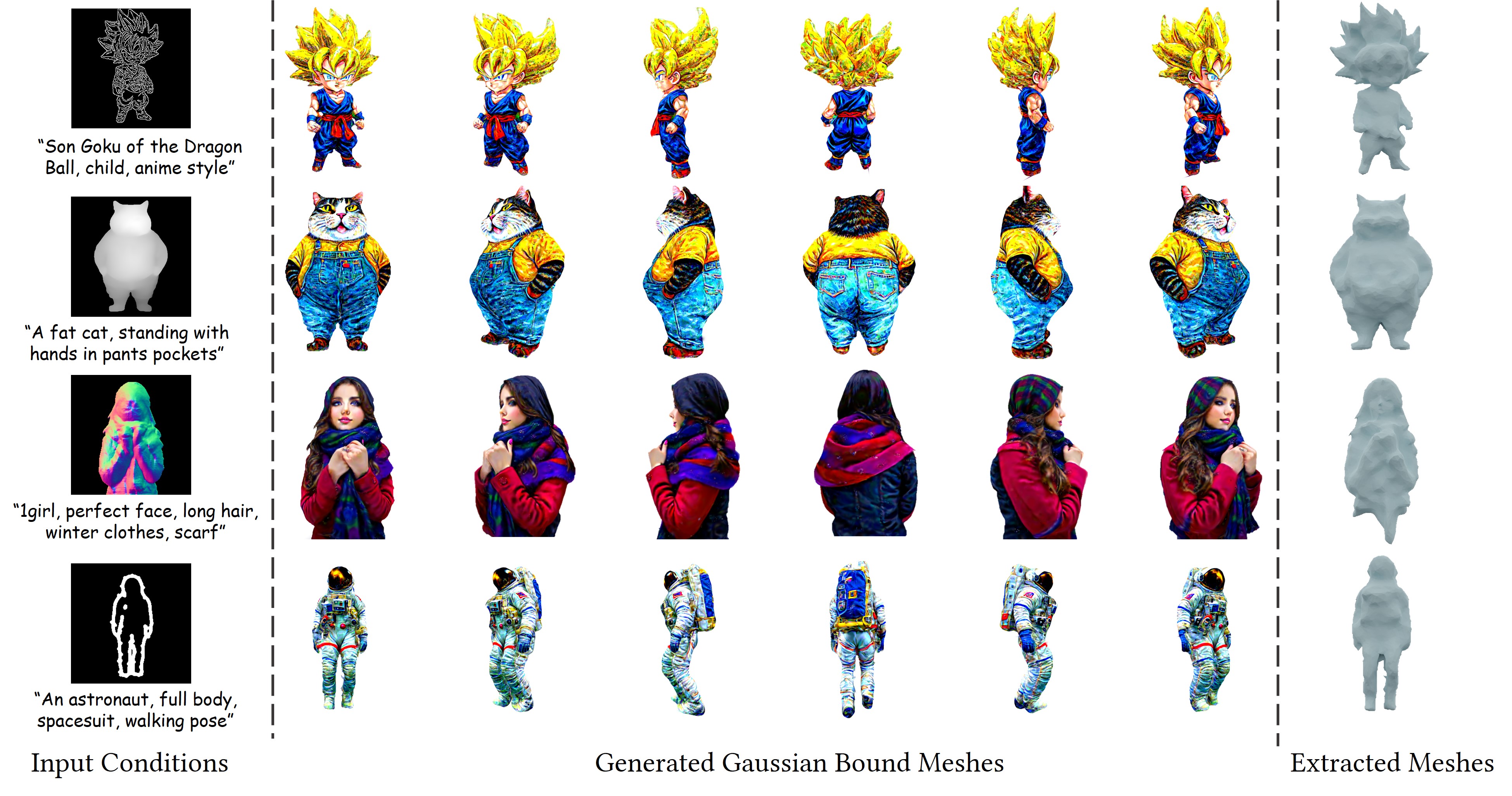}
	\captionof{figure}{Given a text prompt and a condition image, our method is able to achieve high-fidelity and efficient controllable text-to-3D generation of Gaussian binded mesh and textured mesh.}
        \vspace{-1em}
\end{figure*}

\begin{abstract}
While text-to-3D and image-to-3D generation tasks have received considerable attention, one important but under-explored field between them is controllable text-to-3D generation, which we mainly focus on in this work. To address this task, \textbf{1)} we introduce Multi-view ControlNet (MVControl), a novel neural network architecture designed to enhance existing pre-trained multi-view diffusion models by integrating additional input conditions, such as edge, depth, normal, and scribble maps. Our innovation lies in the introduction of a conditioning module that controls the base diffusion model using both local and global embeddings, which are computed from the input condition images and camera poses. Once trained, MVControl is able to offer 3D diffusion guidance for optimization-based 3D generation. And, \textbf{2)} we propose an efficient multi-stage 3D generation pipeline that leverages the benefits of recent large reconstruction models and score distillation algorithm. Building upon our MVControl architecture, we employ a unique hybrid diffusion guidance method to direct the optimization process. In pursuit of efficiency, we adopt 3D Gaussians as our representation instead of the commonly used implicit representations. We also pioneer the use of SuGaR, a hybrid representation that binds Gaussians to mesh triangle faces. This approach alleviates the issue of poor geometry in 3D Gaussians and enables the direct sculpting of fine-grained geometry on the mesh. Extensive experiments demonstrate that our method achieves robust generalization and enables the controllable generation of high-quality 3D content. The source code is available at our website: \url{https://lizhiqi49.github.io/MVControl}.

\end{abstract}    
\section{Introduction}
\label{sec:intro}
%


Remarkable progress has recently been achieved in the field of 2D image generation, which has subsequently propelled research in 3D generation tasks. This progress is attributed to the favorable properties of image diffusion models \cite{rombach2022stablediffusion, liu2023zero123} and differentiable 3D representations \cite{mildenhall2021nerf, wang2021neus, shen2021dmtet, kerbl20233dgaussian}. In particular, recent methods based on score distillation optimization (SDS) \cite{poole2022dreamfusion} have attempted to distill 3D knowledge from pre-trained large text-to-image generative models \cite{rombach2022stablediffusion, liu2023zero123, shi2023mvdream}, leading to impressive results \cite{poole2022dreamfusion, lin2023magic3d, tang2023make, chen2023fantasia3d, metzer2023latent, wang2023prolificdreamer, tang2023dreamgaussian}.

Several approaches aim to enhance generation quality, such as applying multiple optimization stages \cite{lin2023magic3d, chen2023fantasia3d}, optimizing the diffusion prior with 3D representations simultaneously \cite{wang2023prolificdreamer, sun2023dreamcraft3d}, refining score distillation algorithms \cite{katzir2023nfsd, yu2023csd}, and improving pipeline details \cite{huang2023dreamtime, armandpour2023prepneg, zhu2023hifa}. Another focus is on addressing view-consistency issues by incorporating multi-view knowledge into pre-trained diffusion models \cite{liu2023zero123, shi2023mvdream, liu2023syncdreamer, li2023sweetdreamer, qian2023magic123, long2023wonder3d}. However, achieving high-quality 3D assets often requires a combination of these techniques, which can be time-consuming. To mitigate this, recent work aims to train 3D generation networks to produce assets rapidly \cite{nichol2022pointe, jun2023shap, cao2023large, hong2023lrm, li2023instant3d, wang2023pflrm, tang2024lgm}. While efficient, these methods often produce lower quality and less complex shapes due to limitations in training data.

While many works focus on text- or image-to-3D tasks, an important yet under-explored area lies in controllable text-to-3D generation—a gap that this work aims to address. In this work, we propose a new highly efficient controllable 3D generation pipeline that leverages the advantages of both lines of research mentioned in the previous paragraph. Motivated by the achievements of 2D ControlNet \cite{zhang2023controlnet}, an integral component of Stable-Diffusion \cite{rombach2022stablediffusion}, we propose MVControl, a multi-view variant. Given the critical role of multi-view capabilities in 3D generation, MVControl is designed to extend the success of 2D ControlNet into the multi-view domain. We adopt MVDream \cite{shi2023mvdream}, a newly introduced multi-view diffusion network, as our foundational model. MVControl is subsequently crafted to collaborate with this base model, facilitating controllable text-to-multi-view image generation. Similar to the approach in \cite{zhang2023controlnet}, we freeze the weights of MVDream and solely focus on training the MVControl component. However, the conditioning mechanism of 2D ControlNet, designed for single image generation, does not readily extend to the multi-view scenario, making it challenging to achieve view-consistency by directly applying its control network to interact with the base model. Additionally, MVDream is trained on an absolute camera system conflicts with the practical need for relative camera poses in our application scenario. To address these challenges, we introduce a simple yet effective conditioning strategy.  

After training MVControl, we can leverage it to establish 3D priors for controllable text-to-3D asset generation.
To address the extended optimization times of SDS-based methods, which can largely be attributed to the utilization of NeRF\cite{mildenhall2021nerf}-based implicit representations, we propose employing a more efficient explicit 3D representation, 3D Gaussian\cite{kerbl20233dgaussian}. Specifically, we propose a multi-stage pipeline for handling textual prompts and condition images: 1) Initially, we employ our MVControl to generate four multi-view images, which are then inputted into LGM\cite{tang2024lgm}, a recently introduced large Gaussian reconstruction model. This step yields a set of coarse 3D Gaussians. 2) Subsequently, the coarse Gaussians undergo optimization using a hybrid diffusion guidance approach, combining our MVControl with a 2D diffusion model. We introduce SuGaR \cite{guedon2023sugar} regularization terms in this stage to improve the Gaussians' geometry. 3) The optimized Gaussians are then transformed into a coarse Gaussian-bound mesh, for further refinement of both texture and geometry. Finally, a high-quality textured mesh is extracted from the refined Gaussian-bound mesh.

In summary, our main contributions are as follows:
\vspace{-0.3em}
\begin{itemize}
	\itemsep0em
    \item We introduce a novel network architecture designed for controllable fine-grain text-to-multi-view image generation. The model is evaluated across various condition types (edge, depth, normal, and scribble), demonstrating its generalization capabilities;
    \item We develop a multi-stage yet efficient 3D generation pipeline that combines the strengths of large reconstruction models and score distillation. This pipeline optimizes a 3D asset from coarse Gaussians to SuGaR, culminating in a mesh. Importantly, we are the first to explore the potential of a Gaussian-Mesh hybrid representation in the realm of 3D generation;
    \item Extensive experimental results showcase the ability of our method to produce high-fidelity multi-view images and 3D assets. These outputs can be precisely controlled using an input condition image and text prompt.
    %
\end{itemize}

\begin{figure*}[!t]
    \centering
    \includegraphics[width=0.95\linewidth]{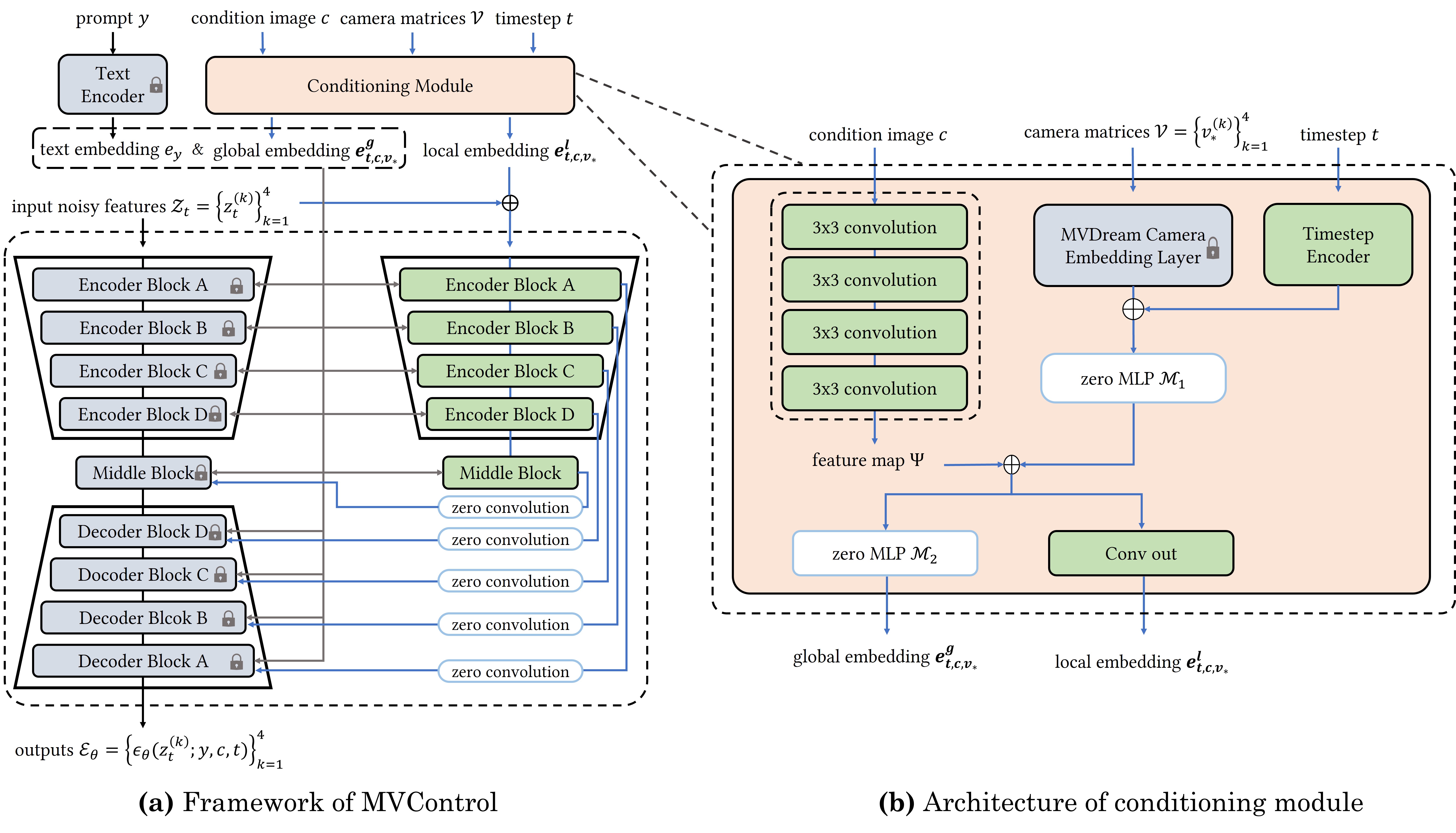}
    \vspace{-1.5em}
    \caption{{\bf{Architecture of proposed MVControl.}} (a) MVControl consists of a frozen multi-view diffusion model and a trainable MVControl. (b) Our model takes care of all input conditions to control the generation process both locally and globally through a conditioning module. (c) Once MVControl is trained, we can exploit it to serve a hybrid diffusion prior for controllable text-to-3D content generation via SDS optimization procedure.}
    \label{fig_mvcontrol}
    \vspace{-1em}
\end{figure*}

\section{Related Work}
\label{sec:related}
\PAR{Multi-view Diffusion Models.} The success of text-to-image generation via large diffusion models inspires the development of multi-view image generation. Commonly adopted approach is to condition on a diffusion model by an additional input image and target pose \cite{liu2023zero123,long2023wonder3d,liu2023syncdreamer}. Unlike those methods, Chan \etal recently proposed to learn 3D scene representation from a single or multiple input images and then exploit a diffusion model for target novel view image synthesis \cite{chan20233diffusion}. Instead of generating a single target view image, MVDiffusion \cite{tang2023mvdiffusion} proposes to generate multi-view consistent images in one feed-forward pass. They build upon a pre-trained diffusion model to have better generalization capability. MVDream \cite{shi2023mvdream} introduces a method for generating consistent multi-view images from a text prompt. They achieve this by fine-tuning a pre-trained diffusion model using a 3D dataset. The trained model is then utilized as a 3D prior to optimize the 3D representation through Score Distillation Sampling. Similar work ImageDream\cite{wang2023imagedream} substitutes the text condition with an image. While prior works can generate impressive novel/multi-view consistent images, fine-grained control over the generated text-to-multi-view images is still difficult to achieve, as what ControlNet \cite{zhang2023controlnet} has achieved for text-to-image generation. Therefore, we propose a multi-view ControlNet (\ie MVControl) in this work to further advance diffusion-based multi-view image generation. 

\PAR{3D Generation Tasks.} The exploration of generating 3D models can typically be categorized into two approaches. The first is SDS-based optimization method, initially proposed by DreamFusion\cite{poole2022dreamfusion}, which aims to extract knowledge for 3D generation through the utilization of pre-trained large image models. SDS-based method benefits from not requiring expansive 3D datasets and has therefore been extensively explored in subsequent works\cite{lin2023magic3d, chen2023fantasia3d, wang2023prolificdreamer, sun2023dreamcraft3d, tang2023dreamgaussian, zhu2023hifa, yi2023gaussiandreamer, qian2023magic123}. These works provide insights of developing more sophisticated score distillation loss functions \cite{wang2023prolificdreamer, qian2023magic123, sun2023dreamcraft3d}, refining optimization strategies \cite{zhu2023hifa, lin2023magic3d, chen2023fantasia3d, sun2023dreamcraft3d, tang2023dreamgaussian}, and employing better 3D representations \cite{chen2023fantasia3d, wang2023prolificdreamer, sun2023dreamcraft3d, tang2023dreamgaussian, yi2023gaussiandreamer}, thereby further enhancing the quality of the generation. Despite the success achieved by these methods in generating high-fidelity 3D assets, they usually require hours to complete the text-to-3D generation process. On the contrary, feed-forward 3D native methods can produce 3D assets within seconds after training on extensive 3D datasets\cite{deitke2023objaverse}. Researchers have explored various 3D representations to achieve improved results, such as volumetric representation\cite{brock2016generative, gadelha20173d, wu2016learning, li2019synthesizing}, triangular mesh\cite{tan2018cvpr, gao2019tog, pavllo2021iccv, youwang2022eccv}, point cloud\cite{pumarola2020cvpr, achlioptas2018learning}, implicit neural representation\cite{park2019cvpr, mescheder2019cvpr, chen2019cvpr, schwarz2022neurips, chan2022eg3d, wang2023rodin, li2023instant3d, hong2023lrm}, as well as the recent 3D Gaussian\cite{tang2024lgm}. While some methods efficiently generate 3D models that meet input conditions, 3D generative methods, unlike image generative modeling, struggle due to limited 3D training assets. This scarcity hinders their ability to produce high-fidelity and diverse 3D objects. Our method merges both approaches: generating a coarse 3D object with a feed-forward method conditioned on MVControl's output, then refining it using SDS loss for the final representation.

\PAR{Optimization-based Mesh Generation.} 
The current single-stage mesh generation method, such as MeshDiffusion\cite{liu2023meshdiffusion}, struggles to produce high-quality mesh due to its highly complex structures. 
To achieve high grade mesh in both geometry and texture, researchers often turn to multi-stage optimization-based methods\cite{lin2023magic3d, chen2023fantasia3d, sun2023dreamcraft3d}. These methods commonly use non-mesh intermediate representations that are easy to process, before transforming them back into meshes with mesh reconstruction methods, which can consume a long optimization time. DreamGaussian\cite{tang2023dreamgaussian} refer to a more efficient representation, 3D Gaussians, to effectively reduce the training time. However, extracting meshes from millions of unorganized tiny 3D Gaussians remains challenging.  LGM\cite{tang2024lgm} presents a new mesh extraction method for 3D Gaussians but still relies on implicit representation. In contrast, we adopt a fully explicit representation, a hybrid of mesh and 3D Gaussians as proposed by SuGaR\cite{guedon2023sugar}. This approach enables us to achieve high-quality mesh generation within reasonable optimization time.

\begin{figure*}[!htbp]
    \centering
    \includegraphics[width=0.95\textwidth]{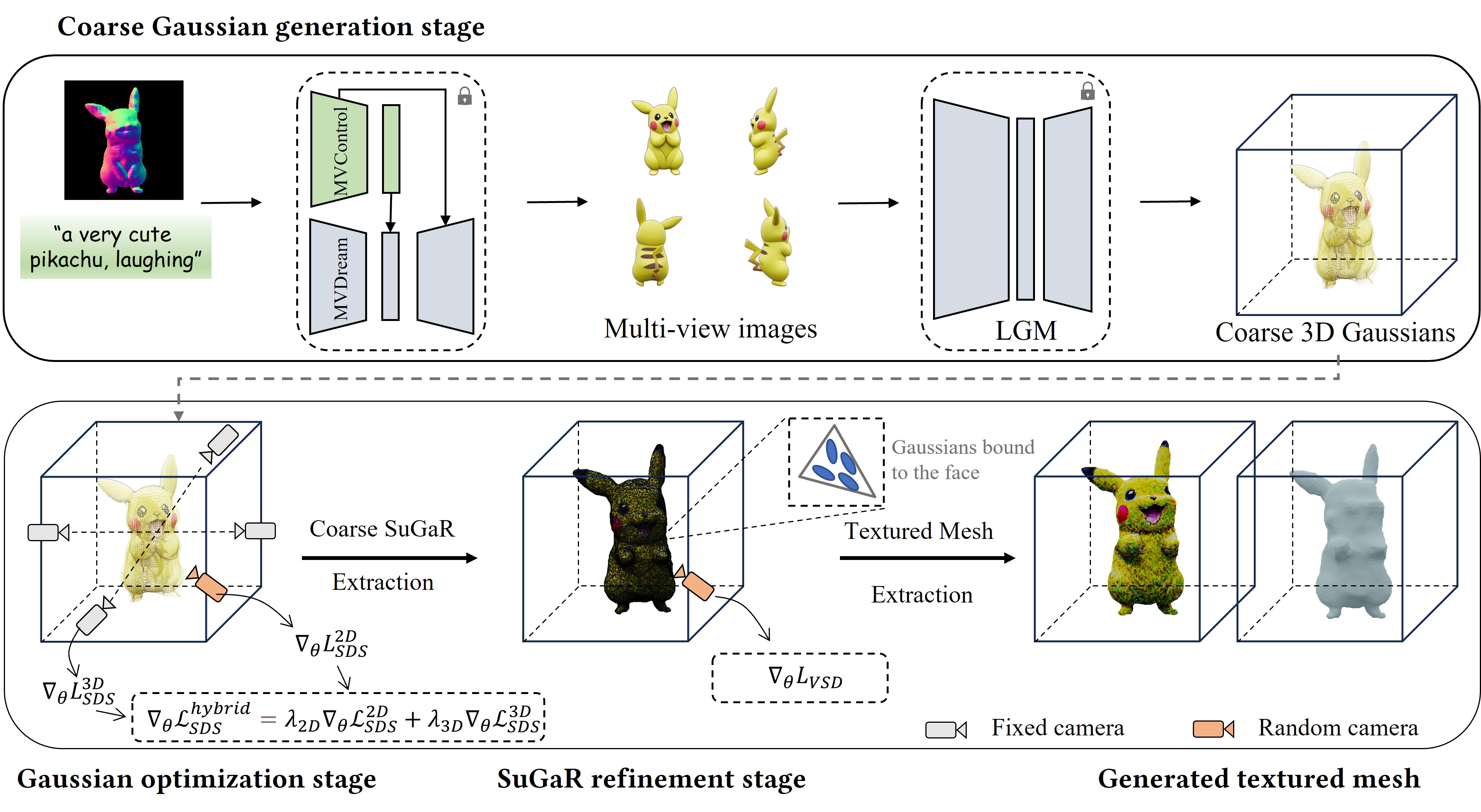}
    \vspace{-1.5em}
    \caption{{\bf{Proposed 3D generation pipeline.}} The multi-stage pipeline can efficiently generate high-quality textured meshes starting from a set of coarse Gaussians generated by LGM, with the input being the multi-view images generated by our MVControl. In the second stage, we employ a 2D \& 3D hybrid diffusion prior for Gaussian optimization. Finally, in the third stage, we calculate the VSD loss to refine the SuGaR representation.}
    \label{fig_3dpipeline}
    \vspace{-1em}
\end{figure*}

\section{Method}
\label{sec:method}

We first review relevant methods, including the 2D ControlNet \cite{zhang2023controlnet}, score distillation sampling \cite{poole2022dreamfusion}, Gaussian Splatting \cite{kerbl20233dgaussian} and SuGaR \cite{guedon2023sugar} in Section \ref{subsec:preliminary}. Then, we analyze the strategy of introducing additional spatial conditioning to MVDream by training a multi-view ControlNet in Section \ref{subsec:MVControl}. Finally in Section \ref{subsec:3dgen}, based on the trained multi-view ControlNet, we propose an efficient 3D generation pipeline, to realize the controllable text-to-3D generation via Gaussian-binded mesh and further textured mesh.

\subsection{Preliminary}
\label{subsec:preliminary}

\PAR{Score Distillation Sampling.} Score distillation sampling (SDS) \cite{poole2022dreamfusion, lin2023magic3d} utilizes a pretrained text-to-image diffusion model as a prior to guide the generation of text-conditioned 3D assets. Specifically, given a pretrained diffusion model $\epsilon_{\phi}$, SDS optimizes the parameters $\theta$ of a differentiable 3D representation (e.g., neural radiance field) using the gradient of the loss $\mathcal{L}_{\textrm{SDS}}$ with respect to $\theta$:
\begin{equation}
\nabla_{\theta}\mathcal{L}_{\textrm{SDS}}(\phi, \mathbf{x})=\mathbb{E}_{t,\epsilon}\left[w(t)(\hat{\epsilon}_\phi-\epsilon)\frac{\partial{z_t}}{\partial{\theta}}\right], \label{eq:3}
\end{equation}
where $\mathbf{x}=g(\theta, c)$ is an image rendered by $g$ under a camera pose $c$, $w(t)$ is a weighting function dependent on the timestep $t$, and $z_t$ is the noisy image input to the diffusion model obtained by adding Gaussian noise $\epsilon$ to $\mathbf{x}$ corresponding to the $t$-th timestep. The primary insight is to enforce the rendered image of the learnable 3D representation to adhere to the distribution of the pretrained diffusion model. In practice, the values of the timestep $t$ and the Gaussian noise $\epsilon$ are randomly sampled at every optimization step.


\PAR{Gaussian Splatting and SuGaR.} Gaussian Splatting \cite{kerbl20233dgaussian} represents the scene as a collection of 3D Gaussians, where each Gaussian $g$ is characterized by its center $\mu_{g}\in \mathbb{R}^3$ and covariance $\Sigma_g\in \mathbb{R}^{3\times3}$. The covariance $\Sigma_g$ is parameterized by a scaling factor $s_g\in \mathbb{R}^3$ and a rotation quaternion $q_g \in \mathbb{R}^4$. Additionally, each Gaussian maintains opacity $\alpha_g\in\mathbb{R}$ and color features $c_g\in\mathbb{R}^C$ for rendering via splatting. Typically, color features are represented using spherical harmonics to model view-dependent effects. During rendering, the 3D Gaussians are projected onto the 2D image plane as 2D Gaussians, and color values are computed through alpha composition of these 2D Gaussians in front-to-back depth order. While the vanilla Gaussian Splatting representation may not perform well in geometry modeling, SuGaR \cite{guedon2023sugar} introduces several regularization terms to enforce flatness and alignment of the 3D Gaussians with the object surface. This facilitates extraction of a mesh from the Gaussians through Poisson reconstruction \cite{kazhdan2006poisson}. Furthermore, SuGaR offers a hybrid representation by binding Gaussians to mesh faces, allowing joint optimization of texture and geometry through backpropagation.

\subsection{Multi-view ControlNet}\label{subsec:MVControl}

Inspired by ControlNet in controlled text-to-image generation and recently released text-to-multi-view image diffusion model (\eg MVDream), we aim to design a multi-view version of ControlNet (\ie MVControl) to achieve controlled text-to-multi-view generation. As shown in \figref{fig_mvcontrol}, we follow similar architecture style as ControlNet, \ie a locked pre-trained MVDream and a trainable control network. The main insight is to preserve the learned prior knowledge of MVDream, while training the control network to learn the inductive bias with a small amount of data. The control network consists of a conditioning module and a copy of the encoder network of MVDream. Our main contribution lies at the conditioning module and we will detail it as follows. 

The conditioning module (\figref{fig_mvcontrol}b) receives the condition image $c$, four camera matrices $\mathcal{V}_*\in\mathbb{R}^{4\times4\times4}$ and timestep $t$ as input, and outputs four local control embeddings $e^l_{t, c, v_*}$ and global control embeddings $e^g_{t, c, v_*}$. The local embedding is then added with the input noisy latent features $\cZ_t\in\mathbb{R}^{4\times C\times H\times W}$ as the input to the control network, and the global embedding $e^g_{t, c, v_*}$ is injected to each layer of MVDream and MVControl to globally control generation.

The condition image $c$ (\ie edge map, depth map \etc) is processed by four convolution layers to obtain a feature map $\Psi$. Instead of using the absolute camera pose matrices embedding of MVDream, we move the embedding into the conditioning module. To make the network better understand the spatial relationship among different views, the relative camera poses $\cV_*$ with respect to the condition image are used. The experimental results also validate the effectiveness of the design. The camera matrices embedding is combined with the timestep embedding, and is then mapped to have the same dimension as the feature map $\Psi$ by a zero-initialized module $\cM_1$. The sum of these two parts is projected to the local embedding $e^l_{t, c, v_*}$ through a convolution layer.

While MVDream is pretrained with absolute camera poses, the conditioning module exploits relative poses as input. We experimentally find that the network hardly converges due to the mismatch of both coordinate frames. We therefore exploit an additional network $\cM_2$ to learn the transformation and output a global embedding $e^g_{t, c, v_*}$ to replace the original camera matrix embedding of MVDream and add on timestep embeddings of both MVDream and MVControl part, so that semantical and view-dependent features are injected globally.

\begin{figure*}[t]
	\centering
	\includegraphics[width=0.95\linewidth]{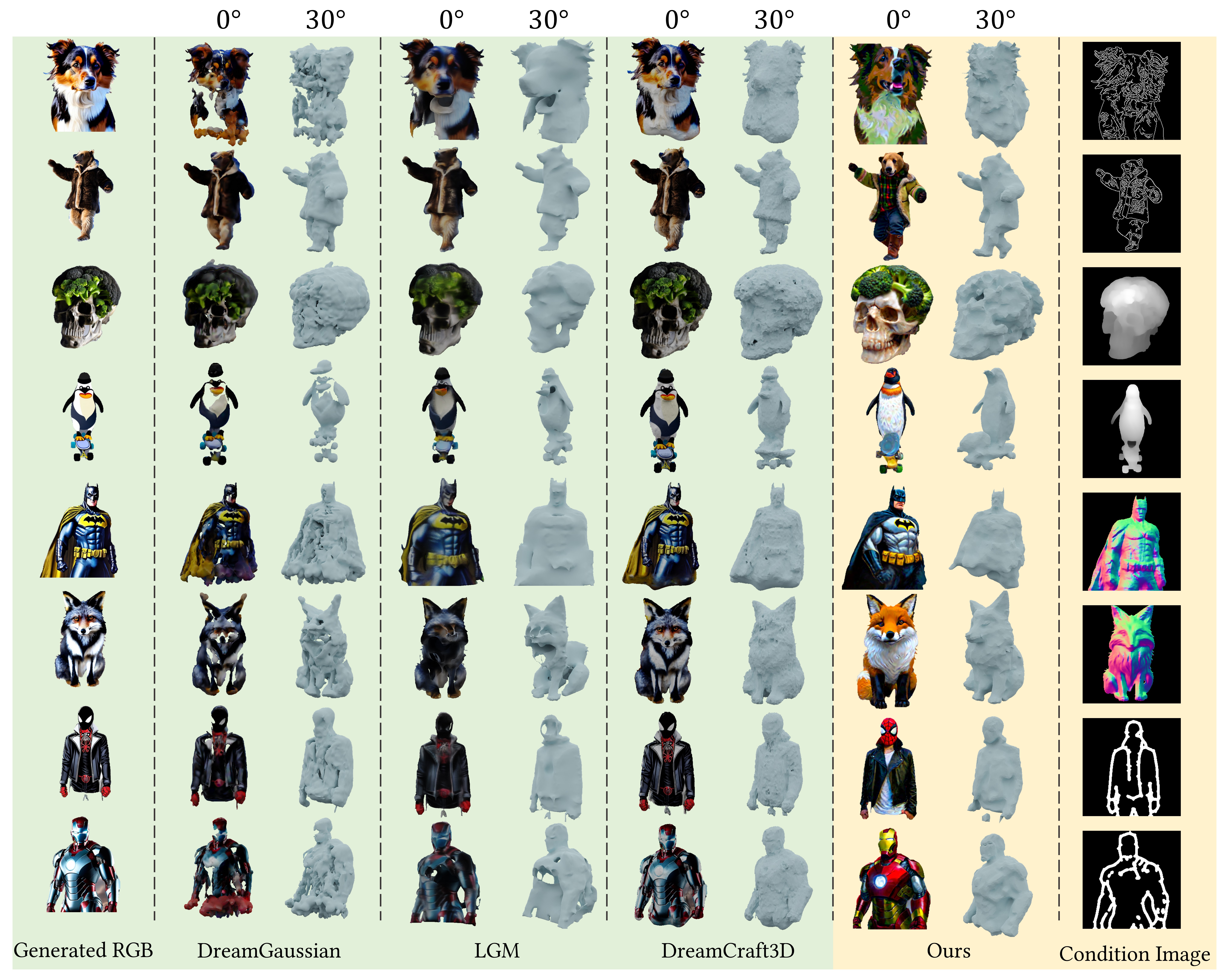}
	\vspace{-1em}
	\captionof{figure}{\textbf{Comparison with baseline 3D generation methods.} Our method yields more delicate texture, and generates much better meshes than the compared methods. We use different color blocks to emphasize that our method only takes the conditioning image rather than RGB as input. Corresponding textual prompts are provided in appendix.}
        \label{fig_3d_comp}
	\vspace{-1em}
\end{figure*}

\subsection{Controllable 3D Textured Mesh Generation}\label{subsec:3dgen}


In this section, we introduce our highly efficient multi-stage textured mesh generation pipeline: Given a condition image and corresponding description prompt, we first generate a set of coarse 3D Gaussians using LGM \cite{tang2024lgm} with four multi-view images generated by our trained MVControl. Subsequently, the coarse Gaussians undergo refinement utilizing a hybrid diffusion prior, supplemented with several regularization terms aimed at enhancing geometry and facilitating coarse SuGaR mesh extraction. Both the texture and geometry of the extracted coarse SuGaR mesh are refined using 2D diffusion guidance under high resolution, culminating in the attainment of a textured mesh. The overall pipeline is illustrated in Fig. \ref{fig_3dpipeline}.

\PAR{Coarse Gaussians Initialization.} Thanks to the remarkable performance of LGM \cite{tang2024lgm}, the images generated by our MVControl model can be directly inputted into LGM to produce a set of 3D Gaussians. However, owing to the low quality of the coarse Gaussians, transferring them directly to mesh, as done in the original paper, does not yield a satisfactory result. Instead, we further apply an optimization stage to refine the coarse Gaussians, with the starting point of optimization either initialized with all the coarse Gaussians' features or solely their positions.


\PAR{Gaussian-to-SuGaR Optimization.} In this stage, we incorporate a hybrid diffusion guidance from a 2D diffusion model and our MVControl to enhance the optimization of coarse Gaussians $\theta$. MVControl offers robust and consistent geometry guidance across four canonical views $\mathcal{V}_*$, while the 2D diffusion model contributes fine geometry and texture sculpting under other randomly sampled views $\mathcal{V}_r\in\mathbb{R}^{B\times4\times4}$. Here, we utilize the DeepFloyd-IF base model \cite{deepfloydif} due to its superior performance in refining coarse geometry.
Given a text prompt $y$ and a condition image $h$, the hybrid SDS gradient $\nabla_\theta\mathcal{L}_{SDS}^{hybrid}$ can be calculated as:
\begin{equation}
\begin{aligned}
    \nabla_\theta\mathcal{L}_{SDS}^{hybrid}=\lambda_{2D} \nabla_\theta\mathcal{L}_{SDS}^{2D}(\mathbf{x}_r=g(\theta, \mathcal{V}_r);t, y)
    \\ + \lambda_{3D} \nabla_\theta\mathcal{L}_{SDS}^{3D}(\mathbf{x}_{*}=g(\theta, \mathcal{V}_*);t, y, h), \label{eq:hybrid_sds}
\end{aligned}
\end{equation}
%
where $\lambda_1$ and $\lambda_2$ are the strength of 2D and 3D prior respectively. 
To enhance the learning of geometry during the Gaussians optimization stage, we employ a Gaussian rasterization engine capable of rendering depth and alpha values \cite{ashawkey2023diffgsrast}. Specifically, in addition to color images, depth $\hat{d}$ and alpha $\hat{m}$ of the scene are also rendered, and we estimate the surface normal $\hat{n}$ by taking the derivative of $\hat{d}$. Consequently, the total variation (TV) regularization terms \cite{rudin1994total} on these components, denoted as $\mathcal{L}_{TV}^{d}$ and $\mathcal{L}_{TV}^{n}$, are calculated and incorporated into the hybrid SDS loss. Furthermore, as the input conditions are invariably derived from existing images, a foreground mask $m_{gt}$ is generated during the intermediate process. Therefore, we compute the mask loss $\mathcal{L}_{mask}=\text{MSE}(\hat{m}, m_{gt})$ to ensure the sparsity of the scene. Thus, the total loss for Gaussian optimization is expressed as:
\begin{equation}
    \mathcal{L}_{GS}=\mathcal{L}_{SDS}^{hybrid} + \lambda_1 \mathcal{L}_{TV}^{d} + \lambda_2 \mathcal{L}_{TV}^{n} + \lambda_3 \mathcal{L}_{mask}, \label{eq:loss_gs}
\end{equation}
where $\lambda_{k={1,2,3}}$ are the weights of depth TV loss, normal TV loss and mask loss respectively. 
Following the approach in \cite{chen2023fantasia3d}, we alternately utilize RGB images or normal maps as input to the diffusion models when calculating SDS gradients. After a certain number of optimization steps $N_1$, we halt the split and pruning of Gaussians. Subsequently, we introduce SuGaR regularization terms \cite{guedon2023sugar} as new loss terms to $\mathcal{L}_{GS}$ to ensure that the Gaussians become flat and aligned with the object surface. This process continues for an additional $N_2$ steps, after which we prune all points whose opacity is below a threshold $\bar{\sigma}$.

\begin{figure}[t]
	\centering
	\includegraphics[width=0.99\linewidth]{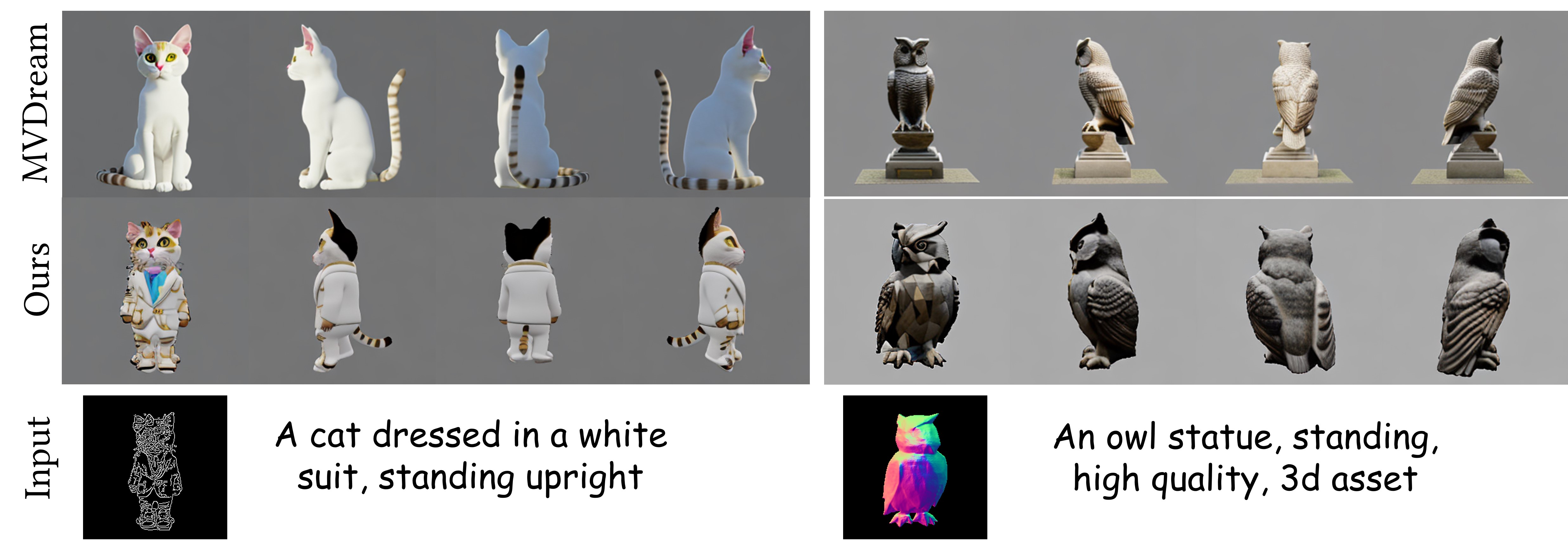}
	\vspace{-0.5em}
	\captionof{figure}{\textbf{Comparison of Multi-view image generation w/ and w/o our MVControl.} MVDream generation results with and without our MVControl attached with edge map and normal map as input condition respectively.}
        \label{fig_2d_comp}
	\vspace{-1em}
\end{figure}

\PAR{SuGaR Refinement.} Following the official pipeline of \cite{guedon2023sugar}, we transfer the optimized Gaussians to a coarse mesh. For each triangle face, a set of new flat Gaussians is bound. The color of these newly bound Gaussians is initialized with the colors of the triangle vertices. The positions of the Gaussians are initialized with predefined barycentric coordinates, and rotations are defined as 2D complex numbers to constrain the Gaussians within the corresponding triangles.
Different from the original implementation, we initialize the learnable opacities of the Gaussians with a large number, specifically 0.9, to facilitate optimization at the outset. Given that the geometry of the coarse mesh is nearly fixed, we replace the hybrid diffusion guidance with solely 2D diffusion guidance computed using Stable Diffusion \cite{rombach2022stablediffusion} to achieve higher optimization resolution. Additionally, we employ Variational Score Distillation (VSD) \cite{wang2023prolificdreamer} due to its superior performance in texture optimization. Similarly, we render the depth $\hat{d}^\prime$ and alpha $\hat{m}^\prime$ through the bound Gaussians. However, in contrast, we can directly render the normal map $\hat{n}^\prime$ using mesh face normals. With these conditions, we calculate the TV losses, $\mathcal{L}_{TV}^{\prime d}$ and $\mathcal{L}_{TV}^{\prime n}$, and the mask loss $\mathcal{L}_{mask}^\prime$ similarly to the previous section. The overall loss for SuGaR refinement is computed as:
\begin{equation}
\mathcal{L}_{SuGaR}=\mathcal{L}_{VSD}+\lambda_1^{\prime}\mathcal{L}_{TV}^{\prime d} + \lambda_2^\prime\mathcal{L}_{TV}^{\prime n} + \lambda_3^\prime\mathcal{L}_{mask}^\prime, \label{eq:loss_sugar}
\end{equation}
where $\lambda^{\prime}_{k={1,2,3}}$ represent the weights of the different loss terms, respectively.

\section{Experiments} \label{sec:exp}

\subsection{Qualitative Comparisons}

\PAR{Multi-view Image Generation.} To assess the controlling capacity of our MVControl, we conduct experiments on MVDream both with and without MVControl attached as shown in \figref{fig_2d_comp}. In the first case, MVDream fails to generate the correct contents according to the given prompt, producing a squatting cat without clothes, which contradicts the prompt. In contrast, it successfully generates the correct contents with the assistance of MVControl. The second case also demonstrates that our MVControl effectively controls the generation of MVDream, resulting in highly view-consistent multi-view images.


\PAR{3D Gaussian-based Mesh Generation.} Given that our 3D generation pipeline aims to produce textured mesh from 3D Gaussians, we compare our method with recent Gaussian-based mesh generation approaches, DreamGaussian \cite{tang2023dreamgaussian} and LGM \cite{tang2024lgm}, both of which can be conditioned on RGB images. Moreover, we also take the state-of-the-art image-to-3D generation method, DreamCraft3D \cite{sun2023dreamcraft3d} into comparison. For fair comparison, we generate 2D RGB images without cherry picking using the pre-trained 2D ControlNet as the input for the compared methods. As illustrated in \figref{fig_3d_comp}, DreamGaussian struggles to generate the geometry for most of the examples, resulting in many broken and hollow areas in the generated meshes. LGM performs better than DreamGaussian, however, its extracted meshes lack details and still contain broken areas in some cases. Although DreamCraft3D can produce unbroken shapes, it still suffer from unsmoothed surface in the meshes. In contrast, our method produces fine-grain meshes with more delicate textures, even without an RGB condition. Due to space limitations, the textual prompts are not provided in \figref{fig_3d_comp}, and we will include them in our appendix. 


%
\begin{table}[t]
    \renewcommand{\arraystretch}{0.95}
    \centering
    \footnotesize
	\setlength\tabcolsep{5pt}
	\begin{tabular}{c|ccc}
		\specialrule{0.1em}{1pt}{1pt}
		               & Optimiztion Stage & CLIP-T$\uparrow$ & CLIP-I$\uparrow$\\
		\specialrule{0.05em}{1pt}{1pt}
            DreamGaussian & GS$\rightarrow$Mesh & 0.200  & 0.847 \\
            LGM           & GS$\rightarrow$Mesh & 0.228  & 0.872 \\
            DreamCraft3D  & NeRF$\rightarrow$NeuS$\rightarrow$DMTet & 0.275  & 0.884 \\
            \specialrule{0.05em}{1pt}{1pt}
            MVControl(Ours)   & GS$\rightarrow$SuGaR & \textbf{0.279}  & \textbf{0.909} \\
		\specialrule{0.1em}{1pt}{1pt}
	\end{tabular}
	\vspace{-1em}
	\caption{{\bf{Quantitative comparison with baselines.}} 
 Our method achieves the best result.
 }
	\label{tab_quant}
	\vspace{-2.2em}
\end{table}

\subsection{Quantitative Comparisons}
In this section, we adopt CLIP-score \cite{mohammad2022clip} to evaluate the compared methods and our method. We calculate both image-text and image-image similarities. For each object, we uniformly render 36 surrounding views. The image-text similarity, denoted as CLIP-T, is computed by averaging the similarities between each view and the given prompt.Similarly, the image-image similarity, referred to as CLIP-I, is the mean similarity between each view and the reference view. The results, calculated for a set of 60 objects, are reported in Table \ref{tab_quant}. When employing our method, the condition type used for each object is randomly sampled from edge, depth, normal, and scribble map. Additionally, the RGB images for DreamGaussian and LGM are generated using 2D ControlNet with the same condition image and prompt. Our method achieves the best performance in terms of both the CLIP-T and CLIP-I score.


\subsection{Ablation Study}

\begin{table}[!h]
    \renewcommand{\arraystretch}{0.9}
	\vspace{-1em}
        \centering
    \footnotesize
	\setlength\tabcolsep{8pt}
	\begin{tabular}{c|ccc}
		\specialrule{0.1em}{1pt}{1pt}
		               & CLIP-T$\uparrow$ & CLIP-I$\uparrow$\\
		\specialrule{0.05em}{1pt}{1pt}
            Stage 2 w/o $\nabla_\theta\mathcal{L}_{SDS}^{3D}$ & 0.245  & 0.866 \\
            Stage 2 w/o normal losses  & 0.263  & 0.876 \\
            Full stage 2 & 0.267  & 0.882 \\
            \specialrule{0.05em}{1pt}{1pt}
            Stage 1 only  & 0.230  & 0.859 \\
            Full method   & \textbf{0.279}  & \textbf{0.909} \\
		\specialrule{0.1em}{1pt}{1pt}
	\end{tabular}
	\vspace{-1em}
	\caption{{\bf{Quantitative comparison of ablation study.}} }
	\label{tab_quant_2}
	\vspace{-1em}
\end{table}

\begin{figure}[!h]
    \centering
    \includegraphics[width=\linewidth]{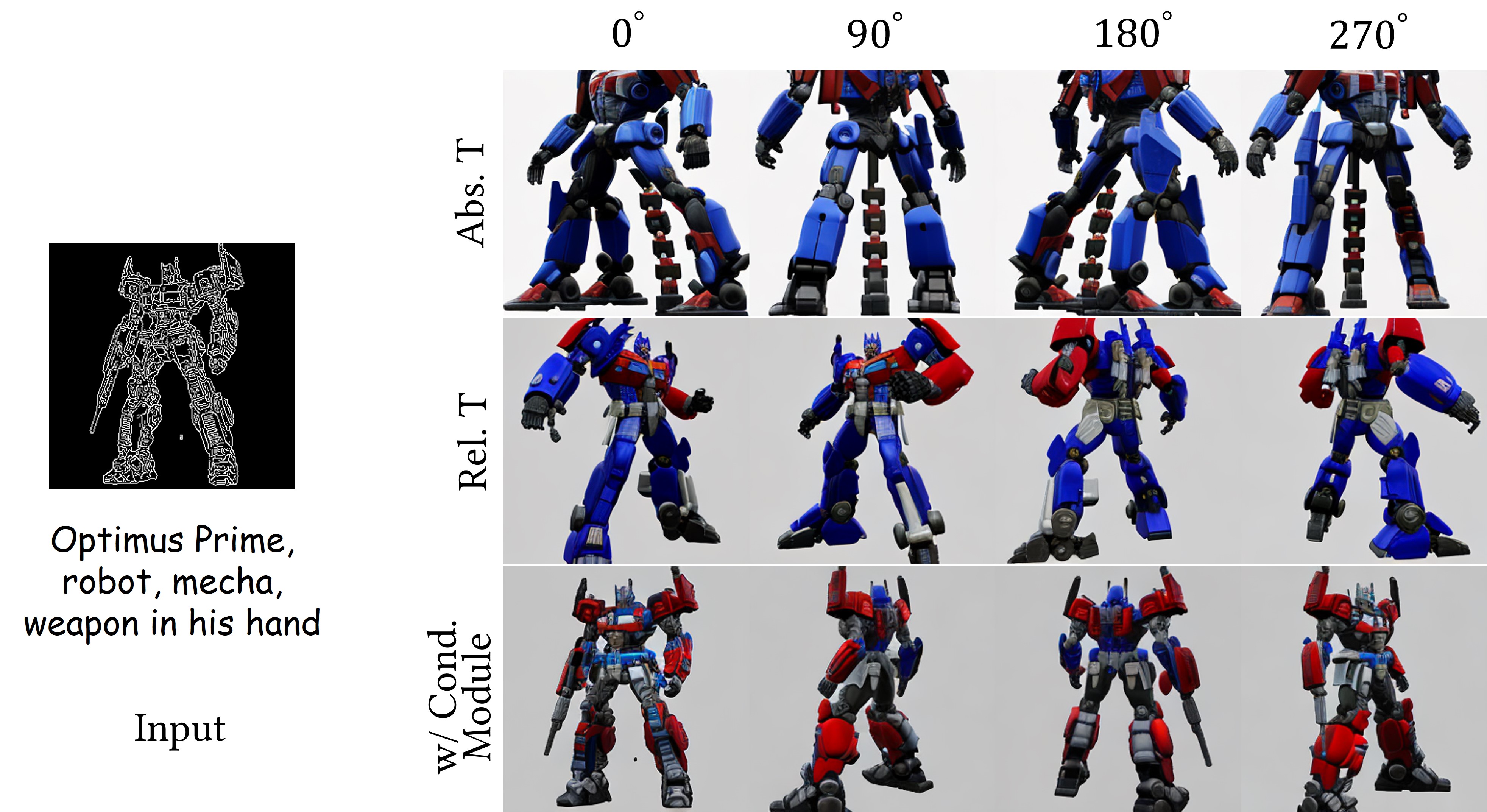}
    \caption{{\bf{Ablation study on conditioning module.}} It achieves precise controlling on multi-view image generation with our conditioning module. }
    \label{fig_ablation_2d}
    \vspace{-1em}
\end{figure}
\PAR{Conditioning Module of MVControl.} We evaluate the training of our model under three different settings to introduce camera condition: 1) we utilize the absolute (world) camera system (i.e., Abs. T) as MVDream \cite{shi2023mvdream} does, without employing our designed conditioning module (retaining the same setup as 2D ControlNet); 2) we adopt the relative camera system without employing the conditioning module; 3) we employ the complete conditioning module. The experimental results, depicted in Fig. \ref{fig_ablation_2d}, demonstrate that only the complete conditioning module can accurately generate view-consistent multi-view images that adhere to the descriptions provided by the condition image. 

\begin{figure}[!h]
    \centering
    \includegraphics[width=\linewidth]{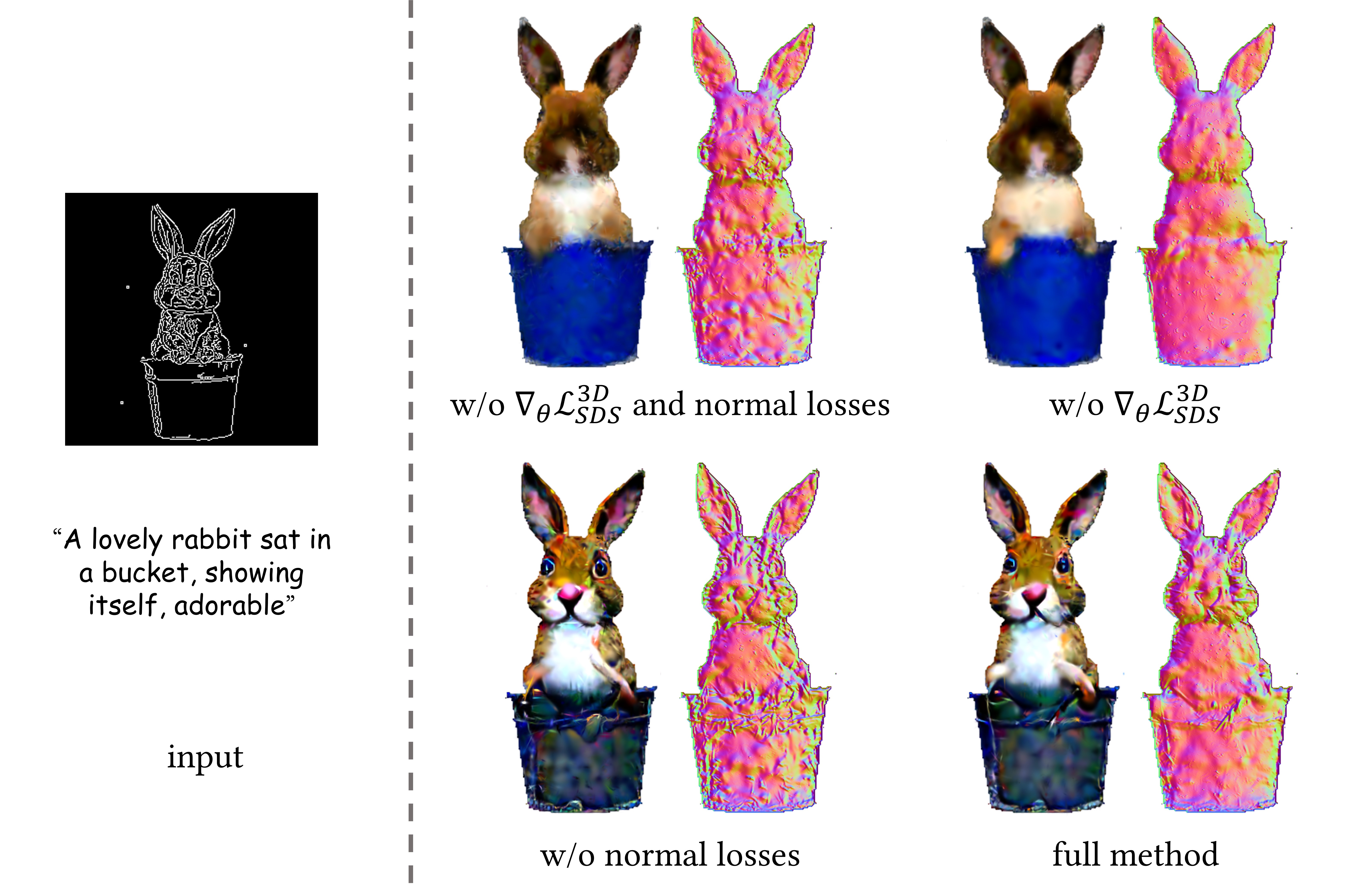}
    \caption{{\bf{Ablation study on Gaussian optimization stage.}} The qualitative results demonstrate the effects of the $\nabla_\theta\mathcal{L}_{SDS}^{3D}$ and normal related loss terms during optimization.}
    \label{fig_ablation_stage2}
    \vspace{-1em}
\end{figure}

\PAR{Hybrid Diffusion Guidance.} We conduct ablation study on hybrid diffusion guidance utilized in the Gaussian optimization stage. As illustrated in Fig. \ref{fig_ablation_stage2} (top right), when excluding $\nabla_\theta \mathcal{L}_{SDS}^{3D}$ provided by our MVControl, the generated 3D Gaussians lack texture details described in the given condition edge map. For instance, the face of the rabbit appears significantly blurrier without $\nabla_\theta \mathcal{L}_{SDS}^{3D}$. The quantitative evaluation is provided in Table \ref{tab_quant_2} (line 1 and 3).

\PAR{Losses on Rendered Normal Maps.} The normal-related losses in our method are alternately calculated using SDS loss with the normal map as input in stage 2, and the normal TV regularization term. We conduct experiments by dropping all of them in stage 2, and the results are illustrated in Fig. \ref{fig_ablation_stage2} (bottom left). Compared to our full method, the surface normal of 3D Gaussians deteriorates without the normal-related losses. The corresponding quantitative results are provided in Table \ref{tab_quant_2} (line 2 and 3).

\begin{figure}[t]
	\centering
	\includegraphics[width=0.99\linewidth]{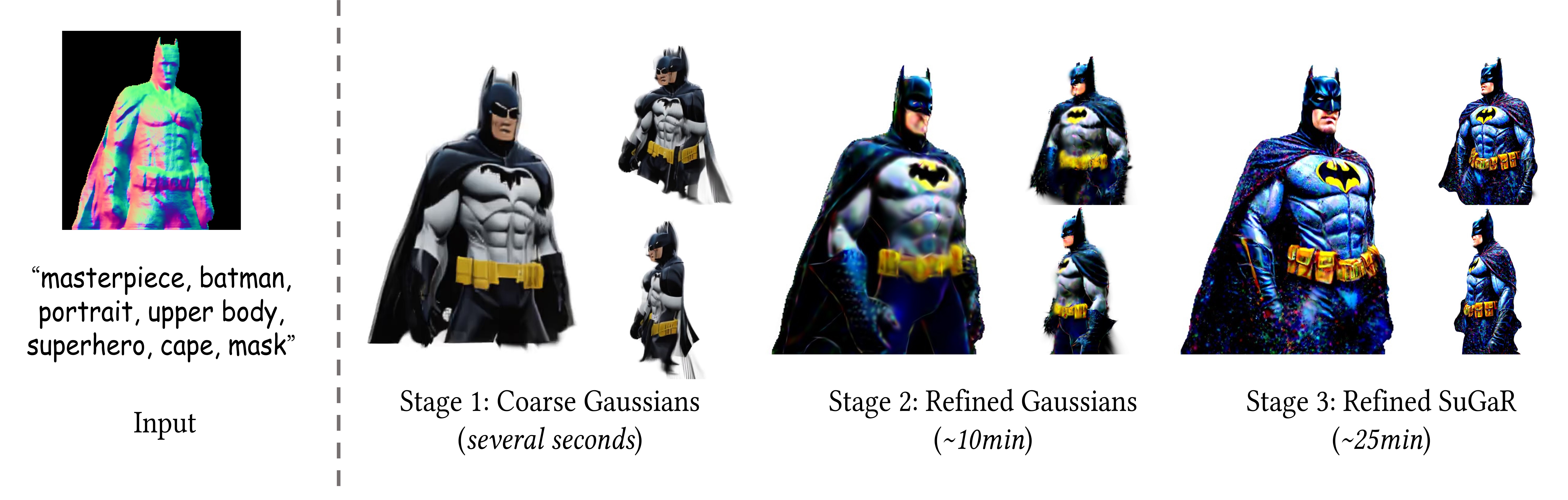}
	\vspace{-0.5em}
	\captionof{figure}{\textbf{Ablation study on multi-stage optimization.} We conduct the ablation study under different optimization stages. Both Gaussian optimization and SuGaR refinement stage promote the quality largely. We also report the consumed time of every stage.}
        \label{fig_ablation_multi_stage}
	\vspace{-1.5em}
\end{figure}

\PAR{Multi-stage Optimization.} We also assess the impact of different optimization stages, as shown in Fig. \ref{fig_ablation_multi_stage}. Initially, in stage 1, the coarse Gaussians exhibit poor geometry consistency. However, after the Gaussian optimization stage, they become view-consistent, albeit with blurry texture. Finally, in the SuGaR refinement stage, the texture of the 3D model becomes fine-grained and of high quality. We have also provided the quantitative evaluation on different optimization stages in Table \ref{tab_quant_2} (the lower 3 lines).



\section{Conclusion}\label{sec:con}
In this work, we delve into the important yet under-explored field of controllable 3D generation. We present a novel network architecture, MVControl, for controllable text-to-multiview image generation. Our approach features a trainable control network that interacts with the base image diffusion model to enable controllable multi-view image generation. Once trained, our network offers 3D diffusion guidance for controllable text-to-3D generation using a hybrid SDS gradient alongside another 2D diffusion model.
We propose an efficient multi-stage 3D generation pipeline using both feed-forward and optimization-based methods. Our pioneering use of SuGaR—an explicit representation blending mesh and 3D Gaussians—outperforms previous Gaussian-based mesh generation approaches. Experimental results demonstrate our method's ability to produce controllable, high-fidelity text-to-multiview images and text-to-3D assets. Furthermore, tests across various conditions show our method's generalization capabilities. We believe our network has broader applications in 3D vision and graphics beyond controllable 3D generation via SDS optimization.

\section*{Acknowledgement}
This work was supported in part by NSFC under Grant 62202389, in part by a grant from the Westlake University-Muyuan Joint Research Institute, and in part by the Westlake Education Foundation. 

{
    \small
    \bibliographystyle{ieeenat_fullname}
    \bibliography{bibliography_zhiqi, bibliography_yiming}
}

\clearpage
\clearpage
\maketitlesupplementary

\appendix
\section{Introduction}
\label{sec:supp_intro}
In this supplementary material, we offer additional details regarding our experimental setup and implementation. Subsequently, we present more qualitative results showcasing the performance and diversity of our method with various types of condition images as input.


\section{Implementation Detail}
\label{sec:supp_impl}

\subsection{Training Data}

\PAR{Multi-view Images Dataset.} 
We employ the multi-view renderings from the publicly available large 3D dataset, Objaverse \cite{deitke2023objaverse}, to train our MVControl. Initially, we preprocess the dataset by removing all samples with a CLIP-score lower than 22, based on the labeling criteria from \cite{sun2023unig3d}. This filtering results in approximately 400k remaining samples.
For each retained sample, we first normalize its scene bounding box to a unit cube centered at the world origin. Subsequently, we sample a random camera setting by uniformly selecting the camera distance between 1.4 and 1.6, the angle of Field-of-View (FoV) between 40 and 60 degrees, the degree of elevation between 0 and 30 degrees, and the starting azimuth angle between 0 and 360 degrees. Under the random camera setting, multi-view images are rendered at a resolution of 256$\times$256 under 4 canonical views at the same elevation starting from the sampled azimuth. We repeat this procedure three times for each object. During training, one of these views is chosen as the reference view corresponding to the condition image. Instead of utilizing the names and tags of the 3D assets, we employ the captions from \cite{luo2023cap3d} as text descriptions for our retained objects. 

\PAR{Canny Edges.}
We apply the Canny edge detector \cite{canny1986computational} with random thresholds to all rendered images to obtain the Canny edge conditions. The lower threshold is randomly selected from the range [50, 125], while the upper threshold is chosen from the range [175, 250].

\PAR{Depth Maps.}
We use the pre-trained depth estimator, Midas \cite{ranftl2021vision}, to estimate the depth maps of rendered images.

\PAR{Normal Maps.}
We compute normal map estimations of all rendered images by computing normal-from-distance on the depth values predicted by Midas.

\PAR{User Scribble.}
We synthesize human scribbles from rendered images by employing an HED boundary detector \cite{xie2015holistically} followed by a set of strong data augmentations, similar to those described in \cite{zhang2023controlnet}.

\subsection{Training Details of MVControl}
While our base model, MVDream \cite{shi2023mvdream}, is fine-tuned from Stable Diffusion v2.1 \cite{rombach2022stablediffusion}, we train our multi-view ControlNet models from publicly available 2D ControlNet checkpoints\footnote{\url{https://huggingface.co/thibaud}} adapted to Stable Diffusion v2.1 for consistency. The models are trained on an 8$\times$A100 node, where we have 160 (40$\times$4) images on each GPU. With a gradient accumulation of 2 steps, we achieve a total batch size of 2560 images. The model undergoes 50000 steps of training under a constant learning rate of $4 \times 10^{-5}$ with 1000 steps of warm-up. Similar to the approach in \cite{zhang2023controlnet}, we randomly drop the text prompt as empty with a 50\% chance during training to facilitate classifier-free learning and enhance the model's understanding of input condition images. Moreover, we also employ 2D-3D joint training following \cite{shi2023mvdream}. Specifically, we randomly sample images from the AES v2 subset of LAION \cite{schuhmann2022laion} with a 30\% probability during training to ensure the network retains its learned 2D image priors. 

\subsection{Implementation Details of 3D Generation}
\PAR{Multi-view Image Generation}. In our coarse Gaussian generation stage, the multi-view images are generated with our MVControl attached to MVDream using a 30-step DDIM sampler \cite{song2020ddim} with a guidance scale of 9 and a negative prompt "ugly, blurry, pixelated obscure, unnatural colors, poor lighting, dull, unclear, cropped, lowres, low quality, artifacts, duplicate". 

\PAR{Gaussian Optimization Stage}. This stage comprises a total of 3000 steps. During the initial 1500 steps, we perform simple 3D Gaussian optimization with split and prune every 300 steps. After step 1500, we cease densification and prune, and instead introduce SuGaR \cite{guedon2023sugar} regularization terms to refine the scene. In the end of the stage, we prune all Gaussians with opacity below $\bar{\sigma}=0.5$. The 3D SDS ($\nabla_{\theta}\mathcal{L}_{SDS}^{3D}$) is computed with a guidance scale of 50 using the CFG rescale trick \cite{lin2023common}, and $\nabla_{\theta}\mathcal{L}_{SDS}^{2D}$ is computed with a guidance scale of 20. We use $\lambda_{2D}=0.1$ and $\lambda_{3D}=0.01$ for 2D and 3D diffusion guidance, with resolutions of 512$\times$512 and 256$\times$256 respectively.

\PAR{SuGaR Refinement Stage}. This stage has totally 5000 steps of optimiztion. The $\nabla_{\theta}\mathcal{L}_{VSD}$ in the SuGaR refinement stage is computed with a guidance scale of 7.5. The training is under resolution 512$\times$512 for rendering.

All score distillation terms also incorporate the aforementioned negative prompt. Our implementation is based on the threestudio project \cite{threestudio2023}. All testing images for condition image extraction are downloaded from \textit{civitai.com}. 

\section{Additional Qualitative Results}
\label{sec:supp_results}

\subsection{Diversity of MVControl}


Similar to 2D ControlNet \cite{zhang2023controlnet}, our MVControl can generate diverse multi-view images with the same condition image and prompt. Please refer to our \href{https://lizhiqi49.github.io/MVControl}{project page} for some of the results.

\subsection{Textured Meshes}





We also provide additional generated textured mesh. 
Please refer to our \href{https://lizhiqi49.github.io/MVControl}{project page} for video and interactive mesh results.

\section{Textual Prompts for 3D Comparison}


Here we provide the missing textual prompts in \figref{fig_3d_comp} of our main paper as below:

1. "RAW photo of A charming long brown coat dog, border collie, head of the dog, upper body, dark brown fur on the back,shelti,light brown fur on the chest,ultra detailed, brown eye"

2. "Wild bear in a sheepskin coat and boots, open-armed, dancing, boots, patterned cotton clothes, cinematic, best quality"

3. "Skull, masterpiece, a human skull made of broccoli"

4. "A cute penguin wearing smoking is riding skateboard, Adorable Character, extremely detailed"

5. "Masterpiece, batman, portrait, upper body, superhero, cape, mask"

6. "Ral-chrome, fox, with brown orange and white fur, seated, full body, adorable"

7. "Spiderman, mask, wearing black leather jacket, punk, absurdres, comic book"

8. "Marvel iron man, heavy armor suit, futuristic, very cool, slightly sideways, portrait, upper body"


\end{document}